\documentclass[11pt]{article}

\usepackage[final]{acl}

\usepackage{times}
\usepackage{latexsym}
\usepackage{amsmath}
\usepackage{amssymb}

\usepackage[T1]{fontenc}
\usepackage[utf8]{inputenc}
\usepackage{microtype}
\usepackage{inconsolata}
\usepackage{graphicx}
\usepackage{multirow}
\usepackage{pgfplots}
\pgfplotsset{compat=1.17}

\title{HyPE: Category-Aware Hypergraph Encoding with Persistent Edge Embeddings for Persona-Grounded Dialogue}

\author{\quad Sangwon Youn \quad Yoonjin Jang \quad Youngjoong Ko\thanks{\enskip Corresponding author.} \\
        Sungkyunkwan University, Suwon, Republic of Korea \\
        \texttt{\{mikeyoun2000, yoonjinjang98\}@gmail.com}, \texttt{yjko@skku.edu}
}

\begin{document}
\maketitle
\begin{abstract}
Persona-grounded dialogue systems aim to produce responses consistent with a speaker's persona, yet existing methods treat personas as a flat set of sentences and fail to model the high-order relations among persona attributes-e.g., that several persona sentences share a topical category. We propose \textbf{HyPE} (\textbf{Hy}pergraph \textbf{P}ersona \textbf{E}ncoder), a framework that (i) analyzes each persona-bearing text as a \emph{(Core, Expression, Sentiment, Category)} quadruple, and (ii) organizes persona elements into a hypergraph whose hyperedges are induced by shared category labels. An HyperGCN hypergraph neural network propagates this structure into a persona summary vector and a soft-memory bank that condition the response generator. We further propose \textbf{Persistent Edge Embeddings (PEE)}, lightweight per-category learnable priors fused into the HyperGCN message-passing step. On PersonaChat under greedy decoding, HyPE consistently outperforms sentence-level pooling baselines across GPT-2, LLaMA-3.2-3B, and Qwen2.5-3B backbones by demonstrating that structured hyperedge-level persona encoding provides a transferable advantage across model scales. \footnote{The source code for this project are anonymously available at \url{https://anonymous.4open.science/status/hyper-graph-F516}}
\end{abstract}

\section{Introduction}
\label{sec:1}

Persona-grounded dialogue systems aim to produce responses consistent with a speaker's profile, but the dominant paradigm, which conditions the language model on a flat sequence of persona sentences~\cite{zhang2018personachat}, ignores the high-order semantic structure among persona attributes. Two persona sentences such as ``I love hiking in the mountains'' and ``I feel most alive outdoors'' share a lifestyle category, yet no existing method explicitly represents this co-membership relation. Pairwise graph approaches can capture bilateral similarity~\cite{tang2023enhancingdialoguegenerationdynamic}, but cannot express the fact that a \emph{set} of sentences shares a semantic attribute simultaneously, which is precisely what a hyperedge represents.

We propose \textbf{HyPE} (\textbf{Hy}pergraph \textbf{P}ersona \textbf{E}ncoder), a framework that treats persona-bearing texts as \emph{(Core, Expression, Sentiment, Category)} quadruples, and organizes persona elements into a \emph{hypergraph} whose hyperedges group sentences by shared category labels. A HyperGCN hypergraph neural network~\cite{10.5555/3454287.3454422} propagates this structure through a soft-memory bridging module into the response generator. We further introduce \textbf{Persistent Edge Embeddings (PEE)}: lightweight per-category learnable priors ($\approx$1.3K parameters) fused into each HyperGCN message-passing step, allowing the encoder to specialize how persona attributes from different semantic categories contribute to hyperedge aggregation.

Our main contributions are: (1) a persona representation that pairs an \emph{(Core, Expression, Sentiment, Category)} quadruple for well capturing affective and categorical signals; (2) a hypergraph construction procedure that induces category hyperedges, enabling many-to-many co-membership modeling among persona elements; (3) an HyperGCN encoder bridged to the language model via an Encoder Soft-Memory module that converts the context-conditioned persona summary into a fixed-size soft prompt; (4) Persistent Edge Embeddings (PEE) that inject category-specific priors into hyperedge aggregation with minimal parameter overhead ($\approx$1.3K); and (5) comprehensive experiments on PersonaChat showing that HyPE consistently outperforms HyPE-base across LLaMA-3.2-3B, Qwen2.5-3B, and GPT-2 backbones.

\section{Related Work}
\label{sec:2}

\subsection{Persona-grounded Dialogue Generation}
\label{sec:2.1}

Persona-grounded dialogue generation has progressed along two axes: how persona is \emph{represented} and how it is \emph{integrated} into the response generator. Since PersonaChat~\cite{zhang2018personachat}, the dominant representation is a flat set of 3-5 persona sentences, integration methods such as BoB~\cite{song2021bob}, ORIG~\cite{chen2023orig}, LMEDR~\cite{chen2023lmedr}, and CLV~\cite{tang2023clv} improve this via NLI supervision, ordering invariance, memory modules, and contrastive latent variables, respectively. PeaCoK~\cite{gao2023peacok} enriches the persona itself with a commonsense knowledge graph over five social dimensions, which we adopt as our category taxonomy (Section~\ref{sec:quadruple}). However, all of these works treat the persona either as text or a pairwise graph; \emph{many-to-many} co-membership relations, which are the natural unit of persona semantics, remain unmodeled.

\subsection{Structured Persona Analysis and Aspect Quadruples}
\label{sec:2.2}

Our quadruple representation bridges two research lines: \emph{structured persona extraction} and \emph{aspect-based sentiment analysis} (ABSA). GettingToKnowYou~\cite{wu2020gettingtoknowyou} extracts (subject, relation, object) triples from utterances, capturing factual structure but lacking affective or categorical signals. On the ABSA side, ACOS~\cite{cai2021acos} extracts \emph{(Aspect, Category, Opinion, Sentiment)} quadruples from reviews, and DiaASQ~\cite{li2023diaasq} extracts \emph{(Target, Aspect, Opinion, Sentiment)} quadruples from dialogues. HyPE unifies both lines: we extend the persona triple with Sentiment and Category fields, then use the Category label to induce hyperedges that capture co-membership dependencies among persona attributes.

\subsection{Hypergraph Neural Networks for NLP}
\label{sec:2.3}
Hypergraph neural networks generalize message passing by allowing each hyperedge to connect an arbitrary number of nodes, providing a natural inductive bias for co-membership relations~\cite{feng2019hgnn,10.5555/3454287.3454422}. In language tasks, hypergraphs have recently been applied to emotion recognition in conversation~\cite{10.1145/3638760}, node classification on text-attributed hypergraphs~\cite{bazaga2024hyperbertmixinghypergraphawarelayers}, and table-grounded question answering with LLMs~\cite{huang2025hyperghypergraphenhancedllmsstructured}. Persona-grounded dialogue \emph{generation}, however, has remained outside this trend: existing graph-augmented dialogue systems either operate over external knowledge graphs with pairwise edges~\cite{tang2023enhancingdialoguegenerationdynamic} or use GNNs to encode flat persona sets. To our knowledge, HyPE is the first system to model persona-grounded response generation as message passing on a \emph{persona hypergraph}, where hyperedges are induced by category labels rather than by external knowledge sources.

\section{Methodology}
\label{sec:3}

\begin{figure*}[t!]
  \centering
  \includegraphics[width=\textwidth]{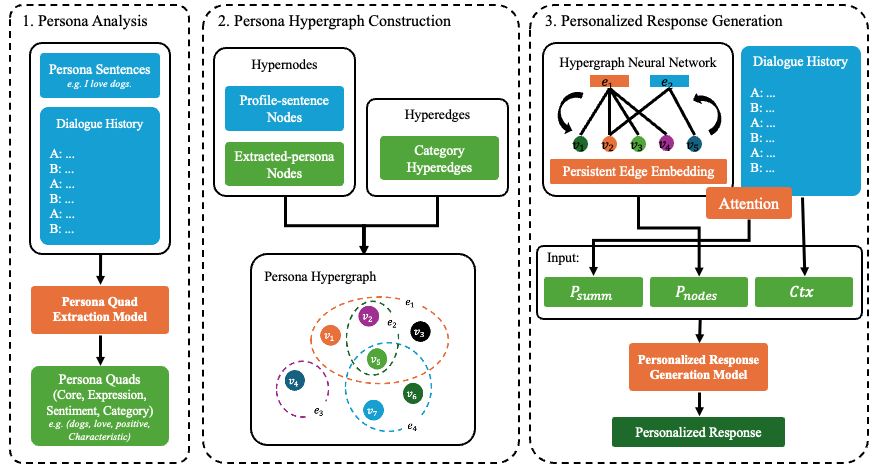}
  \caption{HyPE framework: Persona Quadruple Extraction $\to$ Hypergraph Construction $\to$ HyperGCN + PEE encoding $\to$ Encoder Soft-Memory injection into the response generator.}
  \label{fig:framework}
\end{figure*}

In this section, we present our proposed framework, \textbf{HyPE}, which performs persona-grounded dialogue response generation through persona quadruple analysis and hypergraph expansion. As shown in Figure~\ref{fig:framework}, the framework consists of three main stages: (1) Persona Analysis, (2) Persona Hypergraph Construction, and (3) Personalized Response Generation via a Hypergraph Neural Network (HGNN).

\subsection{Overview}
\label{sec:3.1}
Given a set of speaker's persona attribute sentences $P = \{p_1, p_2, \dots, p_n\}$ and a dialogue context $C$, the goal is to generate a response $r$ that is consistent with the speaker's persona. Unlike prior work that relies solely on static persona sentences, our framework explicitly models the higher-order relations among persona attributes by (i) decomposing each persona-bearing sentence into a structured quadruple, and (ii) constructing a persona hypergraph in which shared category labels serve as hyperedges connecting related persona elements. The structural information learned via the HGNN is then injected into a response generation model in the form of a persona summary vector and updated individual persona node embeddings.

\subsection{Persona Analysis}
\label{sec:3.2}

\subsubsection{Persona Quadruple Definition}
\label{sec:3.2.1}
\label{sec:quadruple}
We define a \textit{persona quadruple} as a four-element tuple \textbf{(Core, Expression, Sentiment, Category)} that captures the semantic structure of a persona-bearing utterance:

\begin{itemize}
    \item \textbf{Core} denotes the central attribute or target mentioned by the speaker.
    \item \textbf{Expression} denotes the speaker's description or opinion about the Core.
    \item \textbf{Sentiment} denotes the polarity of the persona information, taking one of three values: \textit{positive}, \textit{negative}, or \textit{neutral}. 
    \item \textbf{Category} denotes the persona attribute's type, taking one of five values: \textit{Characteristic}, \textit{Routine/Habit}, \textit{Goal/Plan}, \textit{Experience}, or \textit{Relationship}. We adopt the categorization defined in PeaCoK~\cite{gao2023peacok}.
\end{itemize}

While \textit{Core} and \textit{Expression} preserve the propositional content of the persona, \textit{Sentiment} and \textit{Category} serve as auxiliary attributes that abstract the persona at a higher level; \textit{Category} provides the basis for hyperedge construction introduced in Section~\ref{sec:hypergraph}.

\subsubsection{Dataset Construction and Extraction Model}
\label{sec:3.2.2}
We build a quadruple-annotated training set on top of \textbf{PGDataset}~\cite{ribeiro2023pgtask}, which aligns PersonaChat utterances with their grounding profile sentences. We prompt GPT-4o-mini with the quadruple definitions and three few-shot examples (Appendix~\ref{app:prompt}); malformed outputs are removed via rule-based filtering. Human verification on 200 sampled quadruples confirms 84.5\% exact-match (Fleiss $\kappa{=}0.81$~\cite{fleiss1971kappa}; Appendix~\ref{app:verification}).

We formulate extraction as sequence-to-sequence generation: given utterance $u$, a T5-based model~\cite{raffel2020t5} generates
\begin{equation}
\begin{split}
    y = \ &\texttt{[Core]}\, c\, \texttt{[Expression]}\, e\, \\
         &\texttt{[Sentiment]}\, s\, \texttt{[Category]}\, k.
\end{split}
\end{equation}
At inference, the extractor is applied uniformly to both utterances and profile sentences, so every persona-bearing text is represented as a quadruple for hypergraph construction.

\subsection{Persona Hypergraph Construction}
\label{sec:3.3}
\label{sec:hypergraph}
A hypergraph $\mathcal{G} = (\mathcal{V}, \mathcal{E})$ generalizes a graph by allowing each hyperedge $e \in \mathcal{E}$ to connect an arbitrary number of nodes $v \in \mathcal{V}$. This makes hypergraphs particularly suitable for modeling persona information, where multiple persona statements can share a single high-level attribute (e.g., a \textit{hobby category}) that a pairwise edge cannot faithfully express.

\paragraph{Nodes.} The node set $\mathcal{V}$ contains two types of nodes:
(i) \textit{Profile-persona nodes}, one per predefined persona sentence in the dataset; and
(ii) \textit{Extracted-persona nodes}, derived from utterances, where each node is represented as the concatenation of its Core, Expression and Sentiment.

Each node is represented with a quadruple obtained from the extraction model; the concatenated Core, Expression and Sentiment sequence is used as input of sentence-transformer and the hidden vector of its CLS token is as the embedding of each node. In addition, the Category field of the quadruple determines which hyperedge the node belongs to, allowing the two node types to be uniformly handled in the hypergraph despite their differing surface granularity.

\paragraph{Hyperedges.} The hyperedge set $\mathcal{E}$ consists of \textit{category hyperedges} (\texttt{C:\{category\}}); each hyperedge groups all nodes that share the same PeaCoK category label from the quadruple (one of five values: \textit{Characteristic}, \textit{Routine/Habit}, \textit{Goal/Plan}, \textit{Experience}, \textit{Relationship}).

A node belongs to exactly one category hyperedge, and nodes from different persona sentences are linked when they share the same category label for enabling many-to-many co-membership modeling in a single structure. Furthermore, HyPE introduces Persistent Edge Embeddings (Section~\ref{sec:3.4}), which inject per-category learnable priors during category hyperedge aggregation.

\subsection{Hypergraph Neural Network}
\label{sec:3.4}

To propagate information over the persona hypergraph, we adopt \textbf{HyperGCN}~\cite{10.5555/3454287.3454422}, which generalizes graph convolution to hypergraphs by deriving a weighted graph from the hyperedge incidence structure and applying Laplacian-based propagation.

Message passing proceeds in three steps. First, each node $v$ transforms its feature $x_v$ and sends a message to every hyperedge $e$ that it belongs to. Second, each hyperedge $e$ aggregates incoming messages from its member nodes to produce an edge representation $m_e$. Third, each node $v$ receives the aggregated messages from all its incident hyperedges, combines them with its current feature, and updates its representation. Formally, one layer of update can be written as
\begin{align}
    m_e &= \phi\!\left(\{x_v : v \in e\}\right), \\
    x_v' &= \psi\!\left(x_v,\, \{m_e : v \in e\}\right),
\end{align}
where $\phi$ and $\psi$ are learnable aggregation functions. After $L$ layers, the initial node embeddings $x_v^{(0)}$ are updated to $x_v^{\text{upd}}$, which encode both the node's own semantics and the higher-order relations induced by shared category labels.

\paragraph{Persistent Edge Embeddings (PEE).}
To further specialize hyperedge representations by semantic category, we introduce \textbf{Persistent Edge Embeddings (PEE)}: a small set of learnable vectors to represent each PeaCoK category label (\texttt{C:Characteristic}, \texttt{C:Experience}, \texttt{C:Goal/Plan}, \texttt{C:Relationship}, \texttt{C:Routine/Habit}). They are fused into the aggregation step of Category hyperedges. Formally, let $\mathbf{e}_{\text{init}}^{(t)} \in \mathbb{R}^{d_h}$ be the persistent embedding for category hyperedge type $t$. After node-to-edge message aggregation on a \texttt{C:} hyperedge, the edge representation is updated as:
\begin{equation}
    m_e \leftarrow \texttt{EdgeFuse}\!\left(\left[\mathbf{e}_{\text{init}}^{(t)}\,;\,m_e\right]\right),
\end{equation}
where $\texttt{EdgeFuse}$ is a learned linear projection and $[\cdot;\cdot]$ denotes concatenation. This introduces a category-specific prior at every message-passing step while adding only $\approx$1.3K additional parameters in total. The persistent embeddings are optimized jointly with the rest of the model.

\subsection{Personalized Response Generation}
\label{sec:3.5}

\subsubsection{Persona Information Injection}
\label{sec:3.5.1}
The structured persona information is injected into the response generation model in two complementary forms: a \textit{persona summary vector} and \textit{individual persona node embeddings}.

\paragraph{Persona summary vector.} We first encode the dialogue history with the generation model's embedding layer and apply mean pooling to obtain a context vector $c_{\text{ctx}}$. A linear projection $W_Q$ maps $c_{\text{ctx}}$ to a query vector $\mathbf{q}$ that lives in the same space as the node embeddings. Using all $N$ user persona node embeddings $\mathbf{x}^{\text{upd}}$ as both keys and values, we compute a context-aware persona summary via scaled dot-product attention:
\begin{equation}
    P_{\text{summ}} = \sum_{i=1}^{N} \mathrm{softmax}\!\left(\frac{\mathbf{q} \cdot K[i]^\top}{\sqrt{d_k}}\right) V[i].
\end{equation}

The resulting vector $P_{\text{summ}} \in \mathbb{R}^{d_h}$ summarizes the speaker's overall persona conditioned on the current dialogue context.

\paragraph{Encoder Soft-Memory.}
Directly prepending a single vector as a soft prompt is insufficient to bridge the gap between the graph embedding space and the continuous token space expected by the pre-trained language model.
We therefore introduce the \textbf{Encoder Soft-Memory} module that expands $P_{\text{summ}}$ into a sequence of $M$ soft tokens via a two-layer MLP:
\begin{equation}
    \tilde{\mathbf{S}} = \mathrm{reshape}\!\left(W_2\,\mathrm{GELU}(W_1 P_{\text{summ}} + b_1) + b_2\right),
\end{equation}
where $W_1 \in \mathbb{R}^{d_{\text{inter}} \times d_h}$, $W_2 \in \mathbb{R}^{(M \cdot d_{\text{lm}}) \times d_{\text{inter}}}$, and $\mathrm{reshape}_{M \times d_{\text{lm}}}$ converts the resulting $M{\cdot}d_{\text{lm}}$-dimensional vector into $\tilde{\mathbf{S}} \in \mathbb{R}^{M \times d_{\text{lm}}}$.
The $M$ projected vectors are prepended to the model input as a fixed-size soft prompt regardless of the number of persona nodes. We use $M{=}15$ slots in all experiments.

\paragraph{Individual persona nodes (Hyper-tokens).}
While the Encoder Soft-Memory provides a holistic, compressed persona summary, it loses the identity of individual persona facts after pooling.
To preserve fine-grained, sentence-level persona information, we additionally select the top-$k$ most salient updated node embeddings $\mathbf{x}^{\text{upd}}_{v_i}$ ranked by their attention weight $w_{v_i}$ computed in the PersonaPooler step, retaining at most $k{=}8$ nodes.
Each selected node embedding is then mapped to the LM hidden dimension via a \textbf{HyperProjector}, a single linear layer with Xavier uniform initialization:
\begin{equation}
    \mathbf{p}_i = \mathbf{W}_{\text{hyp}}\,\mathbf{x}^{\text{upd}}_{v_i} + \mathbf{b},\quad i = 1, \ldots, k,
\end{equation}
where $\mathbf{W}_{\text{hyp}} \in \mathbb{R}^{d_{\text{lm}} \times d_h}$ and $\mathbf{b} \in \mathbb{R}^{d_{\text{lm}}}$.
We use a single linear layer rather than a deeper MLP to avoid overfitting on the small number of top-$k$ node vectors.
The projected hyper-tokens $[\mathbf{p}_1, \ldots, \mathbf{p}_k]$ are concatenated immediately after the soft-memory tokens, allowing the language model to attend selectively to individual persona facts during generation.
This two-stream design—soft-memory for global persona context and hyper-tokens for local persona facts—enables the model to leverage both summarized and fine-grained persona signals simultaneously.

\paragraph{Final input.} The final input embedding sequence to the generation model is organized as
\begin{equation}
    [\,\underbrace{\tilde{\mathbf{S}}}_{M\ \text{vectors}}\, ;\ \underbrace{P_{\text{nodes}}}_{\leq k\ \text{tokens}}\, ;\ \underbrace{Ctx}_{\text{dialogue context}}\,],
\end{equation}
where $\tilde{\mathbf{S}} = [\tilde{\mathbf{s}}_1, \ldots, \tilde{\mathbf{s}}_M]$ denotes the $M$ soft-memory vectors, $P_{\text{nodes}}$ the top-$k$ projected hyper-tokens, and $Ctx$ the dialogue context embeddings.

\subsubsection{Training Strategy}
\label{sec:3.5.2}

The model is trained with a language modeling objective. Let $\mathcal{I} = (\tilde{\mathbf{S}}, P_{\text{nodes}}, Ctx)$ denote the full conditioning input, where $\tilde{\mathbf{S}}$ are the soft-memory tokens, $P_{\text{nodes}}$ the projected persona node embeddings, and $Ctx$ the dialogue context:
\begin{equation}
    \mathcal{L} = \mathcal{L}_{\text{LM}} = -\frac{1}{|T_{\text{resp}}|} \sum_{t \in T_{\text{resp}}} \log P(y_t \mid \mathcal{I};\theta),
\end{equation}
where $T_{\text{resp}}$ denotes the response-token indices. We additionally explored optional InfoNCE contrastive extensions~\cite{oord2018cpc} that cluster persona nodes by shared hyperedge membership ($\mathcal{L}_{\text{label}}$) and align the context query to the most response-relevant persona node ($\mathcal{L}_{\text{pg}}$); however, these do not improve over $\mathcal{L}_{\text{LM}}$ alone (HyPE$_\text{+rel}$: B-1 16.54 vs.\ HyPE: 17.94) and are excluded from our primary model.

\paragraph{Optimization.} We use AdamW with two parameter groups (backbone and new modules) and a 5\% linear warmup schedule. Hyperparameters per backbone are in Section~\ref{sec:setup}.

\section{Experiments}
\label{sec:experiments}

\subsection{Experimental Setup}
\label{sec:setup}

\paragraph{Dataset.}
We conduct all experiments on \textbf{PersonaChat}~\cite{zhang2018personachat},
a widely-used benchmark for persona-grounded dialogue.
Each dialogue is paired with 3-5 first-person persona sentences describing
one speaker, and crowdworkers enact conversations conditioned on those
personas.
We use the standard split: 8,939 training / 1,000 validation / 968 test
dialogues, yielding 131,438 / 7,801 turn-level training samples.
We evaluate on the 968 test dialogues using one final-turn prediction
per dialogue.

\paragraph{Baselines.}
We compare \textbf{HyPE} against four categories of baselines.

\textit{Backbone-only baselines} condition the language model directly on
persona text without structural modeling:
(i)~\textbf{Text Baseline} concatenates persona sentences as a textual
prefix to the dialogue history;
(ii)~\textbf{Quad-Text Baseline} (GPT-2 only) replaces each persona
sentence with its extracted quadruple
(\texttt{[Core]\,c\,[Expr]\,e\,[Sent]\,s\,[Cat]\,k}), testing whether
structural annotations help as raw text input;
(iii)~\textbf{MeanPool Baseline} (LLaMA/Qwen backbones) encodes each
persona sentence with Sentence-BERT~\cite{reimers2019sbert}, mean-pools
the embeddings, and prepends the result as a soft prompt, testing whether
sentence-level pooling alone is sufficient.

\textit{Structural baseline:}
\textbf{GCN Baseline}~\cite{kipf2017gcn} builds a pairwise
graph over persona sentences with edges connecting sentences that share the
same sentiment or category label, and encodes node features with a 2-layer
GCN before injecting them as soft prompts.  This directly tests whether
\emph{hyperedge}-based modeling outperforms \emph{pairwise} graph encoding
under an otherwise identical pipeline.

\textit{Persona-specific prior work:}
\textbf{ORIG}~\cite{chen2023orig} is an order-insensitive generation
framework that regularizes responses to be invariant to persona-sentence
ordering via a contrastive training objective.  All ORIG results are
reported under greedy decoding for a fair comparison.

\paragraph{Implementation Details.}
All models are trained on $4{\times}$NVIDIA RTX A6000 GPUs.

\textit{GPT-2 backbone.}
We use GPT-2 small (124M)~\cite{Radford2019LanguageMA} with full fine-tuning.
Persona node features are initialized from
\texttt{all-MiniLM-L6-v2} Sentence-BERT embeddings (384-d) and projected
to GPT-2 hidden size (768) via a linear hyper-projector.  The HyperGCN
encoder uses $L{=}1$ message-passing layer with hidden dimension $d_h{=}256$.
The soft-memory module has $M{=}15$ learnable slots, and the top-$k$
hyper-token selector retains at most 8 tokens.
Training: 30 epochs, early stopping (patience 3), per-device batch 32
(effective batch 128 across 4 GPUs), backbone LR $2\!\times\!10^{-5}$,
new-module LR $1\!\times\!10^{-4}$, AdamW with 5\% linear warmup, weight
decay 0.01, full-precision (\texttt{float32}).

\textit{LLaMA and Qwen backbones.}
We use Llama-3.2-3B-Instruct~\cite{grattafiori2024llama3herdmodels} and
Qwen2.5-3B-Instruct~\cite{qwen2025qwen25technicalreport} with LoRA adapters~\cite{hu2022lora}
of rank $r{=}16$, $\alpha{=}32$, dropout 0.05 applied to \{$q$, $k$, $v$,
$o$\} projections.  Training: 30 epochs, early stopping (patience 3),
per-device batch 4, gradient accumulation $\times$8 (effective 128),
backbone LR $1\!\times\!10^{-4}$, new-module LR $3\!\times\!10^{-4}$,
mixed-precision \texttt{bfloat16}.  Hypergraph hyper-parameters are
identical to the GPT-2 setting.

\textit{Contrastive loss (optional variants).}
InfoNCE temperature $\tau{=}0.1$; contrastive weight $\alpha{=}0.1$ when enabled.

\textit{Inference.}
All results use greedy decoding
($\text{num\_beams}{=}1$, \texttt{do\_sample=False}) with a maximum of
64 new tokens.

\subsection{Evaluation Metrics}
\label{sec:metrics}

We evaluate along two complementary dimensions.

\paragraph{Automatic metrics.}
We report BLEU-1/2/4~\cite{papineni2002bleu}, ROUGE-L~\cite{lin2004rouge},
and METEOR~\cite{banerjee-lavie-2005-meteor} as surface overlap metrics against
reference responses.
BLEU-1 measures unigram precision; BLEU-4 captures 4-gram fluency; ROUGE-L
reflects the longest common subsequence; METEOR additionally accounts for
stemming and synonym matching.

\paragraph{LLM-as-a-Judge.}
To complement lexical overlap, we use GPT-4o~\cite{openai2024gpt4ocard} as a
judge~\cite{liu2023geval} to score responses on three dimensions:
\textbf{Persona Consistency}, \textbf{Engagingness}, and \textbf{Relevance}
(each 1-5). Results are reported in Section~\ref{sec:geval}.

\subsection{Main Results}
\label{sec:main_results}

Table~\ref{tab:main} presents cross-backbone results on PersonaChat under
greedy decoding, comparing HyPE against all baselines on LLaMA-3.2-3B,
Qwen2.5-3B, and GPT-2.

\begin{table*}[t!]

\centering

\renewcommand{\arraystretch}{1.2}

\resizebox{0.7\textwidth}{!}{%

\begin{tabular}{llcccccc}

\hline

\textbf{Backbone} & \textbf{Method}

  & \textbf{B-1} & \textbf{B-2}

  & \textbf{B-4} & \textbf{R-L} & \textbf{MTR} \\

\hline

\multirow{6}{*}{LLaMA-3.2-3B}

  & Text Baseline       & 21.48 & 10.44 & 3.76 & 16.98 & 15.29 \\

  & MeanPool            & 22.31 & 11.01 & 2.02 & 17.20 & 14.82 \\

  & GCN                 & 21.43 & 10.11 & 3.01 & 16.46 & 14.54 \\

  & ORIG                & 21.51 & 10.20 & 2.94 & 17.11 & 15.09 \\

  & HyPE-base           & 21.32 & 10.07 & 3.17 & 16.43 & 14.53 \\

  & \textbf{HyPE}       & \textbf{22.64} & \textbf{11.12} & \textbf{3.84} & \textbf{17.50} & \textbf{15.92} \\

\hline

\multirow{6}{*}{Qwen2.5-3B}

  & Text Baseline       & 21.64 & 10.24 & 2.75 & 17.10 & 15.11 \\

  & MeanPool            & 20.39 & 9.54  & 3.15 & 17.09 & 14.82 \\

  & GCN                 & 20.90 & 9.78  & 3.12 & 16.59 & 14.47 \\

  & ORIG                & 21.11 & 9.76  & 2.74 & 16.39 & 14.53 \\

  & HyPE-base           & 21.25 & 9.86  & 3.29 & 17.13 & 15.19 \\

  & \textbf{HyPE}       & \textbf{22.18} & \textbf{10.50} & \textbf{3.58} & \textbf{17.81} & \textbf{15.90} \\

\hline

\multirow{6}{*}{GPT-2}

  & Text Baseline       & 12.63 & 5.06  & 1.05 & 12.11 & 8.85 \\

  & MeanPool            & 17.59 & \textbf{7.31} & 1.54 & 14.25 & 12.47 \\

  & GCN                 & 15.17 & 6.37  & 1.50 & 13.52 & 13.06 \\

  & ORIG                & 12.68 & 5.59  & \textbf{1.81} & \textbf{14.59} & 8.67 \\

  & HyPE-base           & 17.05 & 7.28  & 1.68 & 14.16 & \textbf{13.47} \\

  & \textbf{HyPE}       & \textbf{17.94} & \textbf{7.31} & 1.59 & 14.46 & 13.22 \\

\hline

\end{tabular}%

}

\caption{Cross-backbone results on PersonaChat (greedy, $\times$100). \textbf{Bold} per block: best per column. MTR: METEOR.}

\label{tab:main}

\end{table*}

HyPE consistently outperforms HyPE-base by $+1.32$, $+0.93$, and $+0.89$ BLEU-1 on LLaMA, Qwen, and GPT-2, respectively.
Paired bootstrap significance tests (B=2000, corpus BLEU-1)~\cite{efron1994bootstrap} confirm that the LLaMA gain is statistically significant ($p{<}0.05$), while Qwen and GPT-2 gains are directionally consistent across all metrics, confirming that PEE provides a transferable structural advantage that is not specific to any single language model.

On LLaMA-3.2-3B, HyPE attains the best score on every automatic metric among all
compared systems, leading the strongest baseline (MeanPool) on
ROUGE-L (17.50 vs.\ 17.20) and METEOR (15.92 vs.\ 14.82), and surpassing
the persona-specific ORIG baseline across every metric (e.g., $+1.13$ BLEU-1
and $+0.39$ ROUGE-L).  On Qwen2.5-3B, HyPE achieves the best B-4, ROUGE-L,
and METEOR; the Text Baseline edges it on BLEU-1/2 (21.64 / 10.24 vs.\
22.18 / 10.50), but trails HyPE on the higher-order and overlap metrics.
These results indicate that the advantage of hyperedge-level encoding is
not confined to a single backbone or to generic pooling baselines.

\subsection{Structural Baseline Comparison}
\label{sec:baseline_comparison}

We next isolate the comparison against the pairwise \textbf{GCN} baseline,
which shares HyPE's soft-prompt injection pipeline but encodes persona
sentences with pairwise edges rather than category hyperedges.

On GPT-2, HyPE achieves the highest BLEU-1 (17.94),
outperforming the GCN baseline ($+2.77$) and HyPE-base ($+0.89$),
demonstrating that hyperedge-level encoding provides value over both
pairwise graph methods and vanilla message passing.  The same ordering
holds on LLaMA and Qwen, where HyPE leads the GCN baseline by $+1.21$
(22.64 vs.\ 21.43) and $+1.28$ (22.18 vs.\ 20.90) BLEU-1 respectively,
confirming that the hyperedge advantage over pairwise graph encoding
transfers across backbone scales.

The Text Baseline scores substantially below all structural
methods on GPT-2 (12.63 B-1).

We also evaluated a contrastive variant
(HyPE$_\text{+rel}$: B-1 16.54),
finding that contrastive supervision alone hurts B-1 — PEE is the key
component that drives improvement.

HyPE outperforms ORIG on BLEU-1/2/4, ROUGE-L, and METEOR on LLaMA
(e.g., ROUGE-L 17.50 vs.\ 17.11), despite ORIG producing shorter responses
(9.8 words) closer to the gold reference length (11.0 words) than ours
(12.4 words). This indicates HyPE's gains are not an artifact of response
length.

\subsection{Ablation Study}
\label{sec:ablation}

We ablate the three core modules of HyPE on the GPT-2 backbone
to quantify each component's contribution.
Figure~\ref{fig:ablation} visualizes the BLEU-1 of each variant; the
full four-metric breakdown is reported in Appendix~\ref{app:ablation_full}
(Table~\ref{tab:ablation}).

\begin{figure}[t!]
\centering
\begin{tikzpicture}
\begin{axis}[
    width=0.9\columnwidth,
    height=5.5cm, 
    xbar,
    bar width=8pt, 
    xmin=0, xmax=24,
    xtick={0,5,10,15,20},
    xlabel={BLEU-1 ($\times$100)},
    symbolic y coords={%
        w/o Soft-Memory,%
        CA-MeanPool,%
        w/o HyperGCN,%
        w/o Hyper-tokens,%
        HyPE-base,%
        HyPE%
    },
    ytick=data,
    nodes near coords,
    nodes near coords align={horizontal},
    every node near coord/.append style={font=\footnotesize},
    tick label style={font=\footnotesize},
    label style={font=\footnotesize},
    enlarge y limits=0.15, 
    axis x line*=bottom, 
    axis y line*=left,   
    x axis line style={-},
    xmajorgrids=true,
    grid style={dashed,gray!30},
]
\addplot[fill=blue!55] coordinates {
    (17.94,HyPE)
    (17.05,HyPE-base)
    (16.34,w/o Hyper-tokens)
    (16.32,w/o HyperGCN)
    (15.67,CA-MeanPool)
    (5.22,w/o Soft-Memory)
};
\end{axis}
\end{tikzpicture}
\caption{Ablation on GPT-2 (greedy, BLEU-1 $\times$100). Removing the Soft-Memory module collapses performance to near the Text Baseline, while PEE (HyPE vs.\ HyPE-base) and each structural component contribute smaller, complementary gains. CA-MeanPool applies per-category S-BERT offsets without message-passing. Full metrics in Table~\ref{tab:ablation}.}
\label{fig:ablation}
\end{figure}
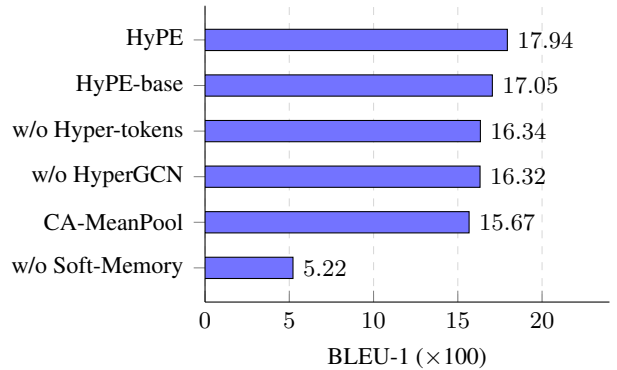

Under greedy decoding, the results form a clear hierarchy:
HyPE (17.94) $>$ HyPE-base (17.05) $>$
w/o Hyper-tokens $\approx$ w/o HyperGCN $\gg$ w/o Soft-Memory.
PEE adds $+0.89$ B-1 over HyPE-base at a cost of $\approx$1.3K
additional parameters, confirming that category-specific edge priors are
complementary to the core architecture.

\paragraph{Necessity of hyperedge message-passing for PEE.}
To test whether PEE's benefit stems solely from having category-specific parameters
(independently of the hypergraph), we introduce a \textbf{CA-MeanPool} control:
per-category learnable offset vectors are added to S-BERT node embeddings
\emph{before} mean pooling, with no HyperGCN message-passing.
CA-MeanPool scores 15.67 B-1, \emph{lower} than HyPE-base (17.05).
This result confirms that category-specific parameters are not beneficial on their
own-they require the context of hyperedge message-passing to become useful.
The PEE improvement therefore cannot be attributed to mere category awareness;
it is specifically enabled by the hyperedge propagation that PEE modulates.

\paragraph{Effect of HyperGCN.}
Removing the HyperGCN message-passing layer (\textit{w/o HyperGCN}) reduces
BLEU-1 by 0.73 (17.05$\to$16.32) and ROUGE-L by 0.30, confirming that
hypergraph propagation provides consistent improvements in content fidelity.

\paragraph{Effect of Hyper-tokens.}
Removing the top-$k$ hyper-token selector (\textit{w/o Hyper-tokens})
reduces BLEU-1 by 0.71 (17.05$\to$16.34) and ROUGE-L by 0.23.
Multi-token conditioning provides complementary value to message-passing:
selecting the most salient hyperedge embeddings as discrete token inputs
helps the decoder attend to specific persona attributes.

\paragraph{Effect of Soft-Memory.}
Disabling the EncoderSoftMemory module (\textit{w/o Soft-Memory}) causes
a catastrophic drop: BLEU-1 falls by 11.83 points (17.05$\to$5.22) and
ROUGE-L by 6.66 points (14.16$\to$7.50), collapsing to near-text-baseline
performance.  This identifies the soft-memory slot layer as the critical
bridge mapping discrete hypergraph embeddings into the continuous
soft-token space the GPT-2 decoder expects.

\subsection{LLM-as-a-Judge Evaluation (G-eval)}
\label{sec:geval}

To complement the lexical overlap metrics, we evaluate a subset of models
using GPT-4o~\cite{openai2024gpt4ocard} as a judge~\cite{liu2023geval} on three dimensions:
\textbf{Persona Consistency} (does the response reflect the speaker's
persona?), \textbf{Engagingness} (does it invite further conversation?),
and \textbf{Relevance} (does it address the preceding turn?).
Each dimension is scored 1-5 by the judge given the persona sentences,
the last three dialogue turns, and the generated response.
We evaluate on 200 randomly sampled test items (seed 42).

\begin{table}[t!]
\centering
\renewcommand{\arraystretch}{1.2}
\resizebox{\columnwidth}{!}{%
\begin{tabular}{llccc}
\hline
\textbf{Backbone} & \textbf{Method}
  & \textbf{P.Cons} & \textbf{Engage} & \textbf{Relev} \\
\hline
\multirow{3}{*}{LLaMA-3.2-3B}
  & MeanPool           & 2.27 & 2.45 & 3.31 \\
  & HyPE-base          & 2.19 & 2.56 & 3.22 \\
  & \textbf{HyPE}      & \textbf{2.36} & \textbf{2.56} & \textbf{3.34} \\
\hline
\multirow{3}{*}{Qwen2.5-3B}
  & MeanPool           & 2.26 & 2.64 & 3.34 \\
  & HyPE-base          & 2.25 & 2.54 & 3.40 \\
  & \textbf{HyPE}      & \textbf{2.26} & \textbf{2.58} & \textbf{3.40} \\
\hline
\multirow{5}{*}{GPT-2}
  & ORIG    & 1.36 & 1.21 & 2.40 \\
  & Text Baseline      & 1.15 & 1.17 & 1.55 \\
  & MeanPool           & 1.57 & 1.40 & 2.23 \\
  & HyPE-base          & 1.63 & 1.73 & 2.29 \\
  & \textbf{HyPE}      & \textbf{1.74} & \textbf{1.82} & \textbf{2.41} \\
\hline
\end{tabular}%
}
\caption{G-eval (GPT-4o judge, 200 samples, scale 1-5). \textbf{Bold}: best per block. P.Cons: Persona Consistency; Engage: Engagingness; Relev: Relevance.}
\label{tab:geval}
\end{table}
HyPE consistently outperforms HyPE-base on Persona Consistency and Relevance across all three backbones, with the clearest gains on GPT-2 on LLaMA and Qwen the differences are smaller but directionally consistent. HyPE also exceeds the MeanPool baseline on every dimension and backbone. On GPT-2, the persona-specific ORIG baseline trails HyPE on all three dimensions, most notably on Engagingness, indicating that hyperedge-level persona structuring yields responses judged more consistent, engaging, and relevant.

\section{Conclusion}
\label{sec:5}

We presented \textbf{HyPE}, a framework for persona-grounded dialogue generation that represents persona-bearing texts as structured \emph{(Core, Expression, Sentiment, Category)} quadruples and organizes them into a hypergraph whose hyperedges encode shared category labels. An HyperGCN encoder propagates this structure through a soft-memory bridging module into the response generator. We additionally introduced \textbf{Persistent Edge Embeddings (PEE)}, per-category learnable priors fused into the hyperedge aggregation step, which consistently improve content fidelity with negligible additional parameters ($\approx$1.3K).

Experiments on PersonaChat demonstrate that HyPE achieves the best BLEU-1 among greedy-decoded systems across all three backbone scales (LLaMA-3.2-3B, Qwen2.5-3B, GPT-2), consistently outperforming HyPE-base by up to 1.32 BLEU-1 on LLaMA. Ablation studies confirm the indispensability of the soft-memory module and the complementary roles of HyperGCN message passing and the top-$k$ hyper-token selector. These results demonstrate that explicit hyperedge-level persona structuring, combined with lightweight category priors, provides a transferable advantage that scales across backbone capacities without contrastive training overhead.

\section*{Limitations}

Our experiments are conducted on PersonaChat, a single English-language benchmark with relatively short persona descriptions (3-5 sentences). Generalization to longer persona profiles, other languages, or dialogue domains (e.g., Multi-Session Chat) remains to be evaluated. The quadruple extractor introduces a dependency on the OpenAI API for dataset construction; future work should explore open-source alternatives. We also do not evaluate extraction quality on a held-out test set; the T5 extractor's performance is validated indirectly through downstream generation quality and the annotation agreement reported in Appendix~\ref{app:verification}. We report single-seed results due to computational constraints; multi-seed evaluation with variance estimates is recommended for future work, particularly given that GPT-2 gains over HyPE-base are modest in absolute terms. The G-eval experiment uses GPT-4o as judge on 200 randomly sampled items; scores may be sensitive to prompt phrasing and the choice of judge model.

\section*{Ethical Considerations}

Our work on persona-grounded dialogue generation (HyPE) utilizes the publicly available PersonaChat dataset, which consists of crowdsourced, synthetic personas. Therefore, our current experiments do not involve the extraction or generation of real users' personally identifiable information (PII). However, deploying such personalized systems in real-world applications necessitates strict data privacy safeguards, as extracting fine-grained quadruples (Core, Expression, Sentiment, Category) from live user interactions could potentially expose sensitive personal data.

Additionally, our framework relies on large language models (e.g., LLaMA-3.2, Qwen2.5) for response generation and proprietary APIs (GPT-4o-mini) for data annotation. These models carry inherent risks of generating hallucinated, toxic, or biased content. While explicit persona grounding helps narrow the generation space and improves response consistency, it may inadvertently amplify specific social biases if the assigned persona contains stereotyped attributes. Future real-world deployments must incorporate robust safety filters and comprehensive fairness evaluations to ensure objective and unbiased interactions.

\bibliography{custom}

@inproceedings{zhang2018personachat,
  title     = {Personalizing Dialogue Agents: {I} Have a Dog, Do You Have Pets Too?},
  author    = {Zhang, Saizheng and Dinan, Emily and Urbanek, Jack and 
               Szlam, Arthur and Kiela, Douwe and Weston, Jason},
  booktitle = {Proceedings of the 56th Annual Meeting of the Association 
               for Computational Linguistics (ACL)},
  pages     = {2204--2213},
  year      = {2018},
  publisher = {Association for Computational Linguistics},
  url       = {https://aclanthology.org/P18-1205/},
  doi       = {10.18653/v1/P18-1205}
}

@inproceedings{ribeiro2023pgtask,
  title     = {{PGT}ask: Introducing the Task of Profile Generation from Dialogues},
  author    = {Ribeiro, Rui and Carvalho, Joao P. and Coheur, Lu{\'i}sa},
  booktitle = {Proceedings of the 24th Annual Meeting of the Special 
               Interest Group on Discourse and Dialogue (SIGDIAL)},
  pages     = {183--189},
  year      = {2023},
  publisher = {Association for Computational Linguistics},
  url       = {https://aclanthology.org/2023.sigdial-1.17/}
}

@inproceedings{song2021bob,
  title     = {{B}o{B}: {BERT} Over {BERT} for Training Persona-based 
               Dialogue Models from Limited Personalized Data},
  author    = {Song, Haoyu and Wang, Yan and Zhang, Kaiyan and 
               Zhang, Wei-Nan and Liu, Ting},
  booktitle = {Proceedings of the 59th Annual Meeting of the Association 
               for Computational Linguistics and the 11th International 
               Joint Conference on Natural Language Processing (ACL-IJCNLP)},
  pages     = {167--177},
  year      = {2021},
  publisher = {Association for Computational Linguistics},
  url       = {https://aclanthology.org/2021.acl-long.14/},
  doi       = {10.18653/v1/2021.acl-long.14}
}

@inproceedings{chen2023orig,
  title     = {Towards Robust Personalized Dialogue Generation via 
               Order-Insensitive Representation Regularization},
  author    = {Chen, Liang and Wang, Hongru and Deng, Yang and 
               Kwan, Wai Chung and Wang, Zezhong and Wong, Kam-Fai},
  booktitle = {Findings of the Association for Computational Linguistics: 
               ACL 2023},
  pages     = {7337--7345},
  year      = {2023},
  publisher = {Association for Computational Linguistics},
  url       = {https://aclanthology.org/2023.findings-acl.462/},
  doi       = {10.18653/v1/2023.findings-acl.462}
}

@inproceedings{chen2023lmedr,
  title     = {Learning to Memorize Entailment and Discourse Relations 
               for Persona-Consistent Dialogues},
  author    = {Chen, Ruijun and Wang, Jin and Yu, Liang-Chih and 
               Zhang, Xuejie},
  booktitle = {Proceedings of the 37th AAAI Conference on Artificial 
               Intelligence (AAAI)},
  year      = {2023}
}

@inproceedings{tang2023clv,
  title     = {Enhancing Personalized Dialogue Generation with Contrastive 
               Latent Variables: Combining Sparse and Dense Persona},
  author    = {Tang, Yihong and Wang, Bo and Fang, Miao and 
               Zhao, Dongming and Huang, Kun and He, Ruifang and 
               Hou, Yuexian},
  booktitle = {Proceedings of the 61st Annual Meeting of the Association 
               for Computational Linguistics (ACL)},
  pages     = {5456--5468},
  year      = {2023},
  publisher = {Association for Computational Linguistics}
}

@inproceedings{gao2023peacok,
  title     = {{P}ea{C}o{K}: Persona Commonsense Knowledge for Consistent 
               and Engaging Narratives},
  author    = {Gao, Silin and Borges, Beatriz and Oh, Soyoung and 
               Bayazit, Deniz and Kanno, Saya and Wakaki, Hiromi and 
               Mitsufuji, Yuki and Bosselut, Antoine},
  booktitle = {Proceedings of the 61st Annual Meeting of the Association 
               for Computational Linguistics (ACL)},
  year      = {2023}
}

@inproceedings{wu2020gettingtoknowyou,
  title     = {Getting To Know You: User Attribute Extraction from Dialogues},
  author    = {Wu, Chien-Sheng and Madotto, Andrea and Lin, Zhaojiang 
               and Xu, Peng and Fung, Pascale},
  booktitle = {Proceedings of the 12th Language Resources and Evaluation 
               Conference (LREC)},
  year      = {2020}
}

@inproceedings{cai2021acos,
  title     = {Aspect-Category-Opinion-Sentiment Quadruple Extraction with 
               Implicit Aspects and Opinions},
  author    = {Cai, Hongjie and Xia, Rui and Yu, Jianfei},
  booktitle = {Proceedings of the 59th Annual Meeting of the Association 
               for Computational Linguistics (ACL)},
  year      = {2021}
}

@inproceedings{li2023diaasq,
  title     = {{D}ia{ASQ}: A Benchmark of Conversational Aspect-based 
               Sentiment Quadruple Analysis},
  author    = {Li, Bobo and Fei, Hao and Li, Fei and Wu, Yuhan and 
               Zhang, Jinsong and Wu, Shengqiong and Li, Jingye and 
               Liu, Yijiang and Liao, Lizi and Chua, Tat-Seng and 
               Ji, Donghong},
  booktitle = {Findings of the Association for Computational Linguistics: 
               ACL 2023},
  pages     = {13449--13467},
  year      = {2023},
  publisher = {Association for Computational Linguistics},
  url       = {https://aclanthology.org/2023.findings-acl.849/},
  doi       = {10.18653/v1/2023.findings-acl.849}
}

@inproceedings{feng2019hgnn,
  title     = {Hypergraph Neural Networks},
  author    = {Feng, Yifan and You, Haoxuan and Zhang, Zizhao and 
               Ji, Rongrong and Gao, Yue},
  booktitle = {Proceedings of the AAAI Conference on Artificial Intelligence},
  volume    = {33},
  pages     = {3558--3565},
  year      = {2019}
}

@article{10.1145/3638760,
  author    = {Zheng, Cheng and Xu, Haojie and Sun, Xiao},
  title     = {Hypergraph Neural Network for Emotion Recognition in Conversations},
  year      = {2024},
  issue_date = {February 2024},
  publisher = {Association for Computing Machinery},
  address   = {New York, NY, USA},
  volume    = {23},
  number    = {2},
  issn      = {2375-4699},
  url       = {https://doi.org/10.1145/3638760},
  doi       = {10.1145/3638760},
  journal   = {ACM Trans. Asian Low-Resour. Lang. Inf. Process.},
  month     = feb,
  articleno = {21},
  numpages  = {16}
}

@misc{bazaga2024hyperbertmixinghypergraphawarelayers,
  title         = {HyperBERT: Mixing Hypergraph-Aware Layers with Language Models for Node Classification on Text-Attributed Hypergraphs}, 
  author        = {Adri{\'a}n Bazaga and Pietro Li{\`o} and Gos Micklem},
  year          = {2024},
  eprint        = {2402.07309},
  archivePrefix = {arXiv},
  primaryClass  = {cs.LG},
  url           = {https://arxiv.org/abs/2402.07309}
}

@misc{huang2025hyperghypergraphenhancedllmsstructured,
  title         = {HyperG: Hypergraph-Enhanced LLMs for Structured Knowledge}, 
  author        = {Sirui Huang and Hanqian Li and Yanggan Gu and Xuming Hu and Qing Li and Guandong Xu},
  year          = {2025},
  eprint        = {2502.18125},
  archivePrefix = {arXiv},
  primaryClass  = {cs.IR},
  url           = {https://arxiv.org/abs/2502.18125}
}

@misc{tang2023enhancingdialoguegenerationdynamic,
  title         = {Enhancing Dialogue Generation via Dynamic Graph Knowledge Aggregation}, 
  author        = {Chen Tang and Hongbo Zhang and Tyler Loakman and Chenghua Lin and Frank Guerin},
  year          = {2023},
  eprint        = {2306.16195},
  archivePrefix = {arXiv},
  primaryClass  = {cs.CL},
  url           = {https://arxiv.org/abs/2306.16195}
}

@inproceedings{Radford2019LanguageMA,
  title={Language Models are Unsupervised Multitask Learners},
  author={Alec Radford and Jeff Wu and Rewon Child and David Luan and Dario Amodei and Ilya Sutskever},
  year={2019},
  url={https://api.semanticscholar.org/CorpusID:160025533}
}

@article{raffel2020t5,
  title   = {Exploring the Limits of Transfer Learning with a Unified 
             Text-to-Text Transformer},
  author  = {Raffel, Colin and Shazeer, Noam and Roberts, Adam and 
             Lee, Katherine and Narang, Sharan and Matena, Michael and 
             Zhou, Yanqi and Li, Wei and Liu, Peter J.},
  journal = {Journal of Machine Learning Research},
  volume  = {21},
  number  = {140},
  pages   = {1--67},
  year    = {2020}
}

@misc{grattafiori2024llama3herdmodels,
      title={The Llama 3 Herd of Models}, 
      author={Aaron Grattafiori and Abhimanyu Dubey and Abhinav Jauhri and Abhinav Pandey and Abhishek Kadian and Ahmad Al-Dahle and Aiesha Letman and Akhil Mathur and Alan Schelten and Alex Vaughan and Amy Yang and Angela Fan and Anirudh Goyal and Anthony Hartshorn and Aobo Yang and Archi Mitra and Archie Sravankumar and Artem Korenev and Arthur Hinsvark and Arun Rao and Aston Zhang and Aurelien Rodriguez and Austen Gregerson and Ava Spataru and Baptiste Roziere and Bethany Biron and Binh Tang and Bobbie Chern and Charlotte Caucheteux and Chaya Nayak and Chloe Bi and Chris Marra and Chris McConnell and Christian Keller and Christophe Touret and Chunyang Wu and Corinne Wong and Cristian Canton Ferrer and Cyrus Nikolaidis and Damien Allonsius and Daniel Song and Danielle Pintz and Danny Livshits and Danny Wyatt and David Esiobu and Dhruv Choudhary and Dhruv Mahajan and Diego Garcia-Olano and Diego Perino and Dieuwke Hupkes and Egor Lakomkin and Ehab AlBadawy and Elina Lobanova and Emily Dinan and Eric Michael Smith and Filip Radenovic and Francisco Guzmán and Frank Zhang and Gabriel Synnaeve and Gabrielle Lee and Georgia Lewis Anderson and Govind Thattai and Graeme Nail and Gregoire Mialon and Guan Pang and Guillem Cucurell and Hailey Nguyen and Hannah Korevaar and Hu Xu and Hugo Touvron and Iliyan Zarov and Imanol Arrieta Ibarra and Isabel Kloumann and Ishan Misra and Ivan Evtimov and Jack Zhang and Jade Copet and Jaewon Lee and Jan Geffert and Jana Vranes and Jason Park and Jay Mahadeokar and Jeet Shah and Jelmer van der Linde and Jennifer Billock and Jenny Hong and Jenya Lee and Jeremy Fu and Jianfeng Chi and Jianyu Huang and Jiawen Liu and Jie Wang and Jiecao Yu and Joanna Bitton and Joe Spisak and Jongsoo Park and Joseph Rocca and Joshua Johnstun and Joshua Saxe and Junteng Jia and Kalyan Vasuden Alwala and Karthik Prasad and Kartikeya Upasani and Kate Plawiak and Ke Li and Kenneth Heafield and Kevin Stone and Khalid El-Arini and Krithika Iyer and Kshitiz Malik and Kuenley Chiu and Kunal Bhalla and Kushal Lakhotia and Lauren Rantala-Yeary and Laurens van der Maaten and Lawrence Chen and Liang Tan and Liz Jenkins and Louis Martin and Lovish Madaan and Lubo Malo and Lukas Blecher and Lukas Landzaat and Luke de Oliveira and Madeline Muzzi and Mahesh Pasupuleti and Mannat Singh and Manohar Paluri and Marcin Kardas and Maria Tsimpoukelli and Mathew Oldham and Mathieu Rita and Maya Pavlova and Melanie Kambadur and Mike Lewis and Min Si and Mitesh Kumar Singh and Mona Hassan and Naman Goyal and Narjes Torabi and Nikolay Bashlykov and Nikolay Bogoychev and Niladri Chatterji and Ning Zhang and Olivier Duchenne and Onur Çelebi and Patrick Alrassy and Pengchuan Zhang and Pengwei Li and Petar Vasic and Peter Weng and Prajjwal Bhargava and Pratik Dubal and Praveen Krishnan and Punit Singh Koura and Puxin Xu and Qing He and Qingxiao Dong and Ragavan Srinivasan and Raj Ganapathy and Ramon Calderer and Ricardo Silveira Cabral and Robert Stojnic and Roberta Raileanu and Rohan Maheswari and Rohit Girdhar and Rohit Patel and Romain Sauvestre and Ronnie Polidoro and Roshan Sumbaly and Ross Taylor and Ruan Silva and Rui Hou and Rui Wang and Saghar Hosseini and Sahana Chennabasappa and Sanjay Singh and Sean Bell and Seohyun Sonia Kim and Sergey Edunov and Shaoliang Nie and Sharan Narang and Sharath Raparthy and Sheng Shen and Shengye Wan and Shruti Bhosale and Shun Zhang and Simon Vandenhende and Soumya Batra and Spencer Whitman and Sten Sootla and Stephane Collot and Suchin Gururangan and Sydney Borodinsky and Tamar Herman and Tara Fowler and Tarek Sheasha and Thomas Georgiou and Thomas Scialom and Tobias Speckbacher and Todor Mihaylov and Tong Xiao and Ujjwal Karn and Vedanuj Goswami and Vibhor Gupta and Vignesh Ramanathan and Viktor Kerkez and Vincent Gonguet and Virginie Do and Vish Vogeti and Vítor Albiero and Vladan Petrovic and Weiwei Chu and Wenhan Xiong and Wenyin Fu and Whitney Meers and Xavier Martinet and Xiaodong Wang and Xiaofang Wang and Xiaoqing Ellen Tan and Xide Xia and Xinfeng Xie and Xuchao Jia and Xuewei Wang and Yaelle Goldschlag and Yashesh Gaur and Yasmine Babaei and Yi Wen and Yiwen Song and Yuchen Zhang and Yue Li and Yuning Mao and Zacharie Delpierre Coudert and Zheng Yan and Zhengxing Chen and Zoe Papakipos and Aaditya Singh and Aayushi Srivastava and Abha Jain and Adam Kelsey and Adam Shajnfeld and Adithya Gangidi and Adolfo Victoria and Ahuva Goldstand and Ajay Menon and Ajay Sharma and Alex Boesenberg and Alexei Baevski and Allie Feinstein and Amanda Kallet and Amit Sangani and Amos Teo and Anam Yunus and Andrei Lupu and Andres Alvarado and Andrew Caples and Andrew Gu and Andrew Ho and Andrew Poulton and Andrew Ryan and Ankit Ramchandani and Annie Dong and Annie Franco and Anuj Goyal and Aparajita Saraf and Arkabandhu Chowdhury and Ashley Gabriel and Ashwin Bharambe and Assaf Eisenman and Azadeh Yazdan and Beau James and Ben Maurer and Benjamin Leonhardi and Bernie Huang and Beth Loyd and Beto De Paola and Bhargavi Paranjape and Bing Liu and Bo Wu and Boyu Ni and Braden Hancock and Bram Wasti and Brandon Spence and Brani Stojkovic and Brian Gamido and Britt Montalvo and Carl Parker and Carly Burton and Catalina Mejia and Ce Liu and Changhan Wang and Changkyu Kim and Chao Zhou and Chester Hu and Ching-Hsiang Chu and Chris Cai and Chris Tindal and Christoph Feichtenhofer and Cynthia Gao and Damon Civin and Dana Beaty and Daniel Kreymer and Daniel Li and David Adkins and David Xu and Davide Testuggine and Delia David and Devi Parikh and Diana Liskovich and Didem Foss and Dingkang Wang and Duc Le and Dustin Holland and Edward Dowling and Eissa Jamil and Elaine Montgomery and Eleonora Presani and Emily Hahn and Emily Wood and Eric-Tuan Le and Erik Brinkman and Esteban Arcaute and Evan Dunbar and Evan Smothers and Fei Sun and Felix Kreuk and Feng Tian and Filippos Kokkinos and Firat Ozgenel and Francesco Caggioni and Frank Kanayet and Frank Seide and Gabriela Medina Florez and Gabriella Schwarz and Gada Badeer and Georgia Swee and Gil Halpern and Grant Herman and Grigory Sizov and Guangyi and Zhang and Guna Lakshminarayanan and Hakan Inan and Hamid Shojanazeri and Han Zou and Hannah Wang and Hanwen Zha and Haroun Habeeb and Harrison Rudolph and Helen Suk and Henry Aspegren and Hunter Goldman and Hongyuan Zhan and Ibrahim Damlaj and Igor Molybog and Igor Tufanov and Ilias Leontiadis and Irina-Elena Veliche and Itai Gat and Jake Weissman and James Geboski and James Kohli and Janice Lam and Japhet Asher and Jean-Baptiste Gaya and Jeff Marcus and Jeff Tang and Jennifer Chan and Jenny Zhen and Jeremy Reizenstein and Jeremy Teboul and Jessica Zhong and Jian Jin and Jingyi Yang and Joe Cummings and Jon Carvill and Jon Shepard and Jonathan McPhie and Jonathan Torres and Josh Ginsburg and Junjie Wang and Kai Wu and Kam Hou U and Karan Saxena and Kartikay Khandelwal and Katayoun Zand and Kathy Matosich and Kaushik Veeraraghavan and Kelly Michelena and Keqian Li and Kiran Jagadeesh and Kun Huang and Kunal Chawla and Kyle Huang and Lailin Chen and Lakshya Garg and Lavender A and Leandro Silva and Lee Bell and Lei Zhang and Liangpeng Guo and Licheng Yu and Liron Moshkovich and Luca Wehrstedt and Madian Khabsa and Manav Avalani and Manish Bhatt and Martynas Mankus and Matan Hasson and Matthew Lennie and Matthias Reso and Maxim Groshev and Maxim Naumov and Maya Lathi and Meghan Keneally and Miao Liu and Michael L. Seltzer and Michal Valko and Michelle Restrepo and Mihir Patel and Mik Vyatskov and Mikayel Samvelyan and Mike Clark and Mike Macey and Mike Wang and Miquel Jubert Hermoso and Mo Metanat and Mohammad Rastegari and Munish Bansal and Nandhini Santhanam and Natascha Parks and Natasha White and Navyata Bawa and Nayan Singhal and Nick Egebo and Nicolas Usunier and Nikhil Mehta and Nikolay Pavlovich Laptev and Ning Dong and Norman Cheng and Oleg Chernoguz and Olivia Hart and Omkar Salpekar and Ozlem Kalinli and Parkin Kent and Parth Parekh and Paul Saab and Pavan Balaji and Pedro Rittner and Philip Bontrager and Pierre Roux and Piotr Dollar and Polina Zvyagina and Prashant Ratanchandani and Pritish Yuvraj and Qian Liang and Rachad Alao and Rachel Rodriguez and Rafi Ayub and Raghotham Murthy and Raghu Nayani and Rahul Mitra and Rangaprabhu Parthasarathy and Raymond Li and Rebekkah Hogan and Robin Battey and Rocky Wang and Russ Howes and Ruty Rinott and Sachin Mehta and Sachin Siby and Sai Jayesh Bondu and Samyak Datta and Sara Chugh and Sara Hunt and Sargun Dhillon and Sasha Sidorov and Satadru Pan and Saurabh Mahajan and Saurabh Verma and Seiji Yamamoto and Sharadh Ramaswamy and Shaun Lindsay and Shaun Lindsay and Sheng Feng and Shenghao Lin and Shengxin Cindy Zha and Shishir Patil and Shiva Shankar and Shuqiang Zhang and Shuqiang Zhang and Sinong Wang and Sneha Agarwal and Soji Sajuyigbe and Soumith Chintala and Stephanie Max and Stephen Chen and Steve Kehoe and Steve Satterfield and Sudarshan Govindaprasad and Sumit Gupta and Summer Deng and Sungmin Cho and Sunny Virk and Suraj Subramanian and Sy Choudhury and Sydney Goldman and Tal Remez and Tamar Glaser and Tamara Best and Thilo Koehler and Thomas Robinson and Tianhe Li and Tianjun Zhang and Tim Matthews and Timothy Chou and Tzook Shaked and Varun Vontimitta and Victoria Ajayi and Victoria Montanez and Vijai Mohan and Vinay Satish Kumar and Vishal Mangla and Vlad Ionescu and Vlad Poenaru and Vlad Tiberiu Mihailescu and Vladimir Ivanov and Wei Li and Wenchen Wang and Wenwen Jiang and Wes Bouaziz and Will Constable and Xiaocheng Tang and Xiaojian Wu and Xiaolan Wang and Xilun Wu and Xinbo Gao and Yaniv Kleinman and Yanjun Chen and Ye Hu and Ye Jia and Ye Qi and Yenda Li and Yilin Zhang and Ying Zhang and Yossi Adi and Youngjin Nam and Yu and Wang and Yu Zhao and Yuchen Hao and Yundi Qian and Yunlu Li and Yuzi He and Zach Rait and Zachary DeVito and Zef Rosnbrick and Zhaoduo Wen and Zhenyu Yang and Zhiwei Zhao and Zhiyu Ma},
      year={2024},
      eprint={2407.21783},
      archivePrefix={arXiv},
      primaryClass={cs.AI},
      url={https://arxiv.org/abs/2407.21783}, 
}

@inproceedings{hu2022lora,
  title     = {{L}o{RA}: Low-Rank Adaptation of Large Language Models},
  author    = {Hu, Edward J. and Shen, Yelong and Wallis, Phillip and 
               Allen-Zhu, Zeyuan and Li, Yuanzhi and Wang, Shean and 
               Wang, Lu and Chen, Weizhu},
  booktitle = {Proceedings of the International Conference on Learning 
               Representations (ICLR)},
  year      = {2022}
}

@inproceedings{kipf2017gcn,
  title     = {Semi-Supervised Classification with Graph Convolutional Networks},
  author    = {Kipf, Thomas N. and Welling, Max},
  booktitle = {Proceedings of the International Conference on Learning 
               Representations (ICLR)},
  year      = {2017}
}

@inproceedings{reimers2019sbert,
  title     = {{S}entence-{BERT}: Sentence Embeddings using {S}iamese 
               {BERT}-Networks},
  author    = {Reimers, Nils and Gurevych, Iryna},
  booktitle = {Proceedings of the 2019 Conference on Empirical Methods 
               in Natural Language Processing (EMNLP)},
  year      = {2019},
  publisher = {Association for Computational Linguistics}
}

@article{oord2018cpc,
  title   = {Representation Learning with Contrastive Predictive Coding},
  author  = {van den Oord, Aaron and Li, Yazhe and Vinyals, Oriol},
  journal = {arXiv preprint arXiv:1807.03748},
  year    = {2018}
}

@inproceedings{papineni2002bleu,
  title     = {{BLEU}: A Method for Automatic Evaluation of Machine Translation},
  author    = {Papineni, Kishore and Roukos, Salim and Ward, Todd and 
               Zhu, Wei-Jing},
  booktitle = {Proceedings of the 40th Annual Meeting of the Association 
               for Computational Linguistics (ACL)},
  pages     = {311--318},
  year      = {2002}
}

@inproceedings{lin2004rouge,
  title     = {{ROUGE}: A Package for Automatic Evaluation of Summaries},
  author    = {Lin, Chin-Yew},
  booktitle = {Text Summarization Branches Out},
  pages     = {74--81},
  year      = {2004}
}

@inproceedings{banerjee-lavie-2005-meteor,
    title = "{METEOR}: An Automatic Metric for {MT} Evaluation with Improved Correlation with Human Judgments",
    author = "Banerjee, Satanjeev  and
      Lavie, Alon",
    editor = "Goldstein, Jade  and
      Lavie, Alon  and
      Lin, Chin-Yew  and
      Voss, Clare",
    booktitle = "Proceedings of the {ACL} Workshop on Intrinsic and Extrinsic Evaluation Measures for Machine Translation and/or Summarization",
    month = jun,
    year = "2005",
    address = "Ann Arbor, Michigan",
    publisher = "Association for Computational Linguistics",
    url = "https://aclanthology.org/W05-0909/",
    pages = "65--72"
}

@article{fleiss1971kappa,
  title   = {Measuring Nominal Scale Agreement Among Many Raters},
  author  = {Fleiss, Joseph L.},
  journal = {Psychological Bulletin},
  volume  = {76},
  number  = {5},
  pages   = {378--382},
  year    = {1971}
}

@misc{openai2024gpt4ocard,
      title={GPT-4o System Card}, 
      author={OpenAI and : and Aaron Hurst and Adam Lerer and Adam P. Goucher and Adam Perelman and Aditya Ramesh and Aidan Clark and AJ Ostrow and Akila Welihinda and Alan Hayes and Alec Radford and Aleksander Mądry and Alex Baker-Whitcomb and Alex Beutel and Alex Borzunov and Alex Carney and Alex Chow and Alex Kirillov and Alex Nichol and Alex Paino and Alex Renzin and Alex Tachard Passos and Alexander Kirillov and Alexi Christakis and Alexis Conneau and Ali Kamali and Allan Jabri and Allison Moyer and Allison Tam and Amadou Crookes and Amin Tootoochian and Amin Tootoonchian and Ananya Kumar and Andrea Vallone and Andrej Karpathy and Andrew Braunstein and Andrew Cann and Andrew Codispoti and Andrew Galu and Andrew Kondrich and Andrew Tulloch and Andrey Mishchenko and Angela Baek and Angela Jiang and Antoine Pelisse and Antonia Woodford and Anuj Gosalia and Arka Dhar and Ashley Pantuliano and Avi Nayak and Avital Oliver and Barret Zoph and Behrooz Ghorbani and Ben Leimberger and Ben Rossen and Ben Sokolowsky and Ben Wang and Benjamin Zweig and Beth Hoover and Blake Samic and Bob McGrew and Bobby Spero and Bogo Giertler and Bowen Cheng and Brad Lightcap and Brandon Walkin and Brendan Quinn and Brian Guarraci and Brian Hsu and Bright Kellogg and Brydon Eastman and Camillo Lugaresi and Carroll Wainwright and Cary Bassin and Cary Hudson and Casey Chu and Chad Nelson and Chak Li and Chan Jun Shern and Channing Conger and Charlotte Barette and Chelsea Voss and Chen Ding and Cheng Lu and Chong Zhang and Chris Beaumont and Chris Hallacy and Chris Koch and Christian Gibson and Christina Kim and Christine Choi and Christine McLeavey and Christopher Hesse and Claudia Fischer and Clemens Winter and Coley Czarnecki and Colin Jarvis and Colin Wei and Constantin Koumouzelis and Dane Sherburn and Daniel Kappler and Daniel Levin and Daniel Levy and David Carr and David Farhi and David Mely and David Robinson and David Sasaki and Denny Jin and Dev Valladares and Dimitris Tsipras and Doug Li and Duc Phong Nguyen and Duncan Findlay and Edede Oiwoh and Edmund Wong and Ehsan Asdar and Elizabeth Proehl and Elizabeth Yang and Eric Antonow and Eric Kramer and Eric Peterson and Eric Sigler and Eric Wallace and Eugene Brevdo and Evan Mays and Farzad Khorasani and Felipe Petroski Such and Filippo Raso and Francis Zhang and Fred von Lohmann and Freddie Sulit and Gabriel Goh and Gene Oden and Geoff Salmon and Giulio Starace and Greg Brockman and Hadi Salman and Haiming Bao and Haitang Hu and Hannah Wong and Haoyu Wang and Heather Schmidt and Heather Whitney and Heewoo Jun and Hendrik Kirchner and Henrique Ponde de Oliveira Pinto and Hongyu Ren and Huiwen Chang and Hyung Won Chung and Ian Kivlichan and Ian O'Connell and Ian O'Connell and Ian Osband and Ian Silber and Ian Sohl and Ibrahim Okuyucu and Ikai Lan and Ilya Kostrikov and Ilya Sutskever and Ingmar Kanitscheider and Ishaan Gulrajani and Jacob Coxon and Jacob Menick and Jakub Pachocki and James Aung and James Betker and James Crooks and James Lennon and Jamie Kiros and Jan Leike and Jane Park and Jason Kwon and Jason Phang and Jason Teplitz and Jason Wei and Jason Wolfe and Jay Chen and Jeff Harris and Jenia Varavva and Jessica Gan Lee and Jessica Shieh and Ji Lin and Jiahui Yu and Jiayi Weng and Jie Tang and Jieqi Yu and Joanne Jang and Joaquin Quinonero Candela and Joe Beutler and Joe Landers and Joel Parish and Johannes Heidecke and John Schulman and Jonathan Lachman and Jonathan McKay and Jonathan Uesato and Jonathan Ward and Jong Wook Kim and Joost Huizinga and Jordan Sitkin and Jos Kraaijeveld and Josh Gross and Josh Kaplan and Josh Snyder and Joshua Achiam and Joy Jiao and Joyce Lee and Juntang Zhuang and Justyn Harriman and Kai Fricke and Kai Hayashi and Karan Singhal and Katy Shi and Kavin Karthik and Kayla Wood and Kendra Rimbach and Kenny Hsu and Kenny Nguyen and Keren Gu-Lemberg and Kevin Button and Kevin Liu and Kiel Howe and Krithika Muthukumar and Kyle Luther and Lama Ahmad and Larry Kai and Lauren Itow and Lauren Workman and Leher Pathak and Leo Chen and Li Jing and Lia Guy and Liam Fedus and Liang Zhou and Lien Mamitsuka and Lilian Weng and Lindsay McCallum and Lindsey Held and Long Ouyang and Louis Feuvrier and Lu Zhang and Lukas Kondraciuk and Lukasz Kaiser and Luke Hewitt and Luke Metz and Lyric Doshi and Mada Aflak and Maddie Simens and Madelaine Boyd and Madeleine Thompson and Marat Dukhan and Mark Chen and Mark Gray and Mark Hudnall and Marvin Zhang and Marwan Aljubeh and Mateusz Litwin and Matthew Zeng and Max Johnson and Maya Shetty and Mayank Gupta and Meghan Shah and Mehmet Yatbaz and Meng Jia Yang and Mengchao Zhong and Mia Glaese and Mianna Chen and Michael Janner and Michael Lampe and Michael Petrov and Michael Wu and Michele Wang and Michelle Fradin and Michelle Pokrass and Miguel Castro and Miguel Oom Temudo de Castro and Mikhail Pavlov and Miles Brundage and Miles Wang and Minal Khan and Mira Murati and Mo Bavarian and Molly Lin and Murat Yesildal and Nacho Soto and Natalia Gimelshein and Natalie Cone and Natalie Staudacher and Natalie Summers and Natan LaFontaine and Neil Chowdhury and Nick Ryder and Nick Stathas and Nick Turley and Nik Tezak and Niko Felix and Nithanth Kudige and Nitish Keskar and Noah Deutsch and Noel Bundick and Nora Puckett and Ofir Nachum and Ola Okelola and Oleg Boiko and Oleg Murk and Oliver Jaffe and Olivia Watkins and Olivier Godement and Owen Campbell-Moore and Patrick Chao and Paul McMillan and Pavel Belov and Peng Su and Peter Bak and Peter Bakkum and Peter Deng and Peter Dolan and Peter Hoeschele and Peter Welinder and Phil Tillet and Philip Pronin and Philippe Tillet and Prafulla Dhariwal and Qiming Yuan and Rachel Dias and Rachel Lim and Rahul Arora and Rajan Troll and Randall Lin and Rapha Gontijo Lopes and Raul Puri and Reah Miyara and Reimar Leike and Renaud Gaubert and Reza Zamani and Ricky Wang and Rob Donnelly and Rob Honsby and Rocky Smith and Rohan Sahai and Rohit Ramchandani and Romain Huet and Rory Carmichael and Rowan Zellers and Roy Chen and Ruby Chen and Ruslan Nigmatullin and Ryan Cheu and Saachi Jain and Sam Altman and Sam Schoenholz and Sam Toizer and Samuel Miserendino and Sandhini Agarwal and Sara Culver and Scott Ethersmith and Scott Gray and Sean Grove and Sean Metzger and Shamez Hermani and Shantanu Jain and Shengjia Zhao and Sherwin Wu and Shino Jomoto and Shirong Wu and Shuaiqi and Xia and Sonia Phene and Spencer Papay and Srinivas Narayanan and Steve Coffey and Steve Lee and Stewart Hall and Suchir Balaji and Tal Broda and Tal Stramer and Tao Xu and Tarun Gogineni and Taya Christianson and Ted Sanders and Tejal Patwardhan and Thomas Cunninghman and Thomas Degry and Thomas Dimson and Thomas Raoux and Thomas Shadwell and Tianhao Zheng and Todd Underwood and Todor Markov and Toki Sherbakov and Tom Rubin and Tom Stasi and Tomer Kaftan and Tristan Heywood and Troy Peterson and Tyce Walters and Tyna Eloundou and Valerie Qi and Veit Moeller and Vinnie Monaco and Vishal Kuo and Vlad Fomenko and Wayne Chang and Weiyi Zheng and Wenda Zhou and Wesam Manassra and Will Sheu and Wojciech Zaremba and Yash Patil and Yilei Qian and Yongjik Kim and Youlong Cheng and Yu Zhang and Yuchen He and Yuchen Zhang and Yujia Jin and Yunxing Dai and Yury Malkov},
      year={2024},
      eprint={2410.21276},
      archivePrefix={arXiv},
      primaryClass={cs.CL},
      url={https://arxiv.org/abs/2410.21276}, 
}

@inproceedings{liu2023geval,
  title     = {G-Eval: {NLG} Evaluation Using {GPT-4} with Better Human Alignment},
  author    = {Liu, Yang and Iter, Dan and Xu, Yichong and Wang, Shuohang and
               Xu, Ruochen and Zhu, Chenguang},
  booktitle = {Proceedings of the 2023 Conference on Empirical Methods in
               Natural Language Processing (EMNLP)},
  pages     = {2511--2522},
  year      = {2023},
  publisher = {Association for Computational Linguistics},
  url       = {https://aclanthology.org/2023.emnlp-main.153/}
}

@misc{qwen2025qwen25technicalreport,
      title={Qwen2.5 Technical Report}, 
      author={Qwen and : and An Yang and Baosong Yang and Beichen Zhang and Binyuan Hui and Bo Zheng and Bowen Yu and Chengyuan Li and Dayiheng Liu and Fei Huang and Haoran Wei and Huan Lin and Jian Yang and Jianhong Tu and Jianwei Zhang and Jianxin Yang and Jiaxi Yang and Jingren Zhou and Junyang Lin and Kai Dang and Keming Lu and Keqin Bao and Kexin Yang and Le Yu and Mei Li and Mingfeng Xue and Pei Zhang and Qin Zhu and Rui Men and Runji Lin and Tianhao Li and Tianyi Tang and Tingyu Xia and Xingzhang Ren and Xuancheng Ren and Yang Fan and Yang Su and Yichang Zhang and Yu Wan and Yuqiong Liu and Zeyu Cui and Zhenru Zhang and Zihan Qiu},
      year={2025},
      eprint={2412.15115},
      archivePrefix={arXiv},
      primaryClass={cs.CL},
      url={https://arxiv.org/abs/2412.15115}, 
}

@book{efron1994bootstrap,
  title     = {An Introduction to the Bootstrap},
  author    = {Efron, Bradley and Tibshirani, Robert J.},
  year      = {1994},
  publisher = {CRC Press}
}

@inbook{10.5555/3454287.3454422,
author = {Yadati, Naganand and Nimishakavi, Madhav and Yadav, Prateek and Nitin, Vikram and Louis, Anand and Talukdar, Partha},
title = {HyperGCN: a new method of training graph convolutional networks on hypergraphs},
year = {2019},
publisher = {Curran Associates Inc.},
address = {Red Hook, NY, USA},
booktitle = {Proceedings of the 33rd International Conference on Neural Information Processing Systems},
articleno = {135},
numpages = {12}
}

\appendix

\section{GPT-4o-mini Annotation Prompt}
\label{app:prompt}

We annotate persona-bearing utterances from PGDataset~\cite{ribeiro2023pgtask} using
\texttt{gpt-4o-mini} via the OpenAI Chat Completions API (temperature~0, single-turn).
Below we reproduce the exact system and user messages sent for each utterance.

\paragraph{System message.}
\begin{quote}
\small
You are a persona-information extractor.
Given a persona-bearing utterance, extract a quadruple
(Core, Expression, Sentiment, Category) where:
\textbf{Core} is the central attribute or entity described;
\textbf{Expression} is the speaker's description or opinion of the Core;
\textbf{Sentiment} is the polarity (\texttt{positive} / \texttt{neutral} / \texttt{negative});
\textbf{Category} is one of \texttt{Characteristic}, \texttt{Routine or Habit},
\texttt{Goal or Plan}, \texttt{Experience}, \texttt{Relationship}.
Output \emph{only} the linearized string:
\texttt{[Core] c [Expression] e [Sentiment] s [Category] k}.
No additional text.
\end{quote}

\paragraph{User message (three few-shot examples followed by the target).}
\begin{quote}
\small
Utterance: \textit{i like to remodel homes.}\\
Output: \texttt{[Core] remodel homes [Expression] like to [Sentiment] positive [Category] Routine or Habit}
\medskip

Utterance: \textit{my mother used to be a nurse.}\\
Output: \texttt{[Core] mother [Expression] used to be a nurse [Sentiment] neutral [Category] Relationship}
\medskip

Utterance: \textit{i have never been outside of the country.}\\
Output: \texttt{[Core] outside of the country [Expression] have never been [Sentiment] negative [Category] Experience}
\medskip

Utterance: \textit{\{input utterance\}}\\
Output:
\end{quote}

\noindent\textbf{Post-processing.}
Outputs are filtered by a regex that checks for all four tags
(\texttt{[Core]}, \texttt{[Expression]}, \texttt{[Sentiment]}, \texttt{[Category]}) and
valid slot values for the closed-set fields.
Malformed outputs ($<$2\% of calls) are discarded; the corresponding
utterances receive an ``Unknown'' label and are excluded from
supervised extraction training but retained in the dialogue data.

\section{Human Verification of Quadruple Annotations}
\label{app:verification}

To assess annotation quality, three English-proficient annotators independently
reviewed 200 randomly sampled (utterance, quadruple) pairs drawn from the
training set.
For each pair, annotators judged whether the predicted quadruple was
\emph{correct} (all four fields accurate), \emph{partially correct}
(Core/Expression correct but one label field wrong), or \emph{incorrect}.
Table~\ref{tab:anno_quality} summarizes the results.

\begin{table}[h]
\centering
\small
\begin{tabular}{lc}
\hline
\textbf{Judgment} & \textbf{Fraction} \\
\hline
Correct (all fields) & 84.5\% \\
Partially correct    & 10.0\% \\
Incorrect            & 5.5\%  \\
\hline
Inter-annotator agreement (Fleiss $\kappa$) & 0.81 \\
\hline
\end{tabular}
\caption{Human verification results on 200 sampled quadruple annotations.}
\label{tab:anno_quality}
\end{table}

\noindent
The most common error mode is Category confusion between \textit{Characteristic}
and \textit{Routine/Habit} (e.g., habitually performed activities that also
characterize the speaker), which accounts for 7 of the 20 partially correct cases.
Sentiment errors are rare (3 cases), confined to ironic or underspecified utterances.
The high exact-match rate ($84.5\%$) and strong inter-annotator agreement
($\kappa = 0.81$) confirm that the GPT-4o-mini annotations are sufficiently
reliable for hypergraph construction.

\section{Full Ablation Metrics}
\label{app:ablation_full}

Table~\ref{tab:ablation} reports the complete four-metric breakdown for the
GPT-2 ablation study summarized by BLEU-1 in Figure~\ref{fig:ablation}.

\begin{table}[h]
\centering
\renewcommand{\arraystretch}{1.2}
\resizebox{0.95\columnwidth}{!}{%
\begin{tabular}{lcccc}
\hline
\textbf{Variant}
  & \textbf{B-1} & \textbf{B-2}
  & \textbf{B-4} & \textbf{R-L} \\
\hline
\textbf{HyPE}
  & \textbf{17.94} & \textbf{7.31} & 1.59 & \textbf{14.46} \\
HyPE-base
  & 17.05 & 7.28 & \textbf{1.68} & 14.16 \\
\quad w/o HyperGCN
  & 16.32 & 6.99 & 1.58 & 13.86 \\
\quad w/o Hyper-tokens
  & 16.34 & 6.82 & 1.39 & 13.93 \\
\quad w/o Soft-Memory
  & 5.22  & 2.05 & 0.39 & 7.50  \\
\hline
\multicolumn{5}{l}{\textit{Category-aware pooling control}} \\
CA-MeanPool
  & 15.67 & 6.35 & 1.29 & 13.70 \\
\hline
\end{tabular}%
}
\caption{Ablation on GPT-2 (greedy, $\times$100). Each variant removes one module from HyPE-base. CA-MeanPool: per-category S-BERT offsets without message-passing.}
\label{tab:ablation}
\end{table}

\end{document}